# CQural: A Novel CNN based Hybrid Architecture for Quantum Continual Machine Learning

Sanyam Jain, Indian Institute of Technology, Jodhpur, India 342037

*Abstract*—Training machine learning models in an incremental fashion is not only important but also an efficient way to achieve artificial general intelligence. The ability that humans possess of continuous or lifelong learning helps them to not forget previously learned tasks. However, current neural network models are prone to catastrophic forgetting when it comes to continual learning. Many researchers have come up with several techniques in order to reduce the effect of forgetting from neural networks, however, all techniques are studied classically with a very less focus on changing the machine learning model architecture. In this research paper, we show that it is not only possible to circumvent catastrophic forgetting in continual learning with novel hybrid classical-quantum neural networks, but also ex- plains what features are most important to learn for classification. In addition, we also claim that if the model is trained with these explanations, it tends to give better performance and learn specific features that are far from the decision boundary. Finally, we present the experimental results to show comparisons between classical and classical-quantum hybrid architectures on benchmark MNIST and CIFAR-10 datasets. After successful runs of learning procedure, we found hybrid neural network outperforms classical one in terms of remembering the right evidences of the class-specific features.

## I. Introduction

QUANTUM Machine Learning (QML) has gained exponential exposure after 2019 when Google announced quantum supremacy [1]. Modern-day Machine Learning and Quantum Computing evolved parallelly, however, an interdisciplinary approach has gained much interest to combine Quantum Computing to enhance and speed up classical machine learning to increase the efficiency of the training algorithm.

Quantum Deep Learning (part of QML) has embarked on an interesting place in the research community. Previous work [2] in QML has claimed to gain quadratic and logarithmic speedups for traditional ML algorithms and has also shown to achieve quadratic and logarithmic speedups for traditional ML algorithms. Some existing algorithms such as Quantum Bayesian Inference [3], Quantum Perceptron Algorithm [4], Quantum Boltzmann Machine [5], and Quantum SVM [6]. Modern machine learning models tend to be resource-intensive to find complex decision boundaries over an input dataset. As complexity of the features in the dataset increases, models take more resources to learn those features.

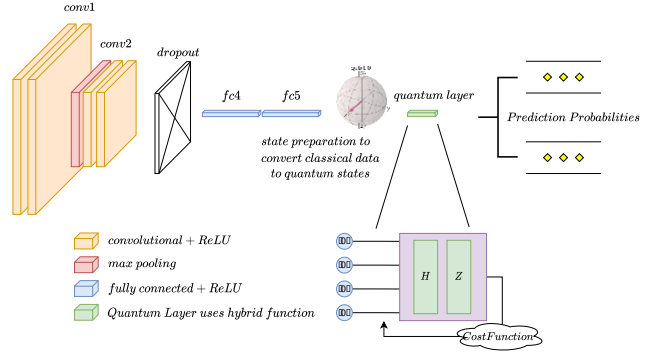

*Figure 1 CQural Architecture*

Continual learning [7] is an advanced incremental online learning theory driven by the idea of lifelong learning. In other words, a learning paradigm is applied for a continuously learning machine learning model. In this particular research, many challenges have to be encountered. One of the prominent and popular defects in continual learning is catastrophic forgetting. Catastrophic forgetting is a tendency of Deep Neural Networks to forget previously learned tasks when they are presented with new tasks. In other words, some examples on which the model is less confident and changes its prediction labels with the course of multiple learning stages are called forgotten examples. Previous research [8] had been done in this line of work tells about the theoretical and statistical proofs that reduce the chances of forgetting events. In addition, a recent work [9] based on calculating the label shifts (label-dispersion) helps investigate the change in labels and prediction confidence for a particular mini-batch of examples—empirical studies on the developments of forgetting resulted in inefficient learning models. Previous work [8] also shows some examples which are not informative and hence forgotten in multiple stages of learning cycles can be eliminated directly. Theoretical claims and empirical approach suggest that examples that are being forgotten are far-off in the feature space from the consistently unforgettable examples. In other words, the examples that are forgotten continuously or their labels shift frequently are good candidates for informativeness. The unforgettable examples and their labels do not shift continuously are less informative. However, this is not true in datasets with fewer examples or with biased data.

A new methodology was proposed in a recent work [10] to prevent catastrophic forgetting in quantum continual learning. Theoretically, it solves the problem by detecting a geometric

S. Jain is student with the Interdisciplinary Research Programme in Quantum Information & Computation, Indian Institute of Technology, Jodhpur, India, 342037. E-mail: sanyam.1@iitj.ac.in

shift in the feature space of the second dataset once the learner is trained on the first dataset. It is expected that there shall be no rise in forgetting of labels when a new dataset is introduced. In recent work, [11], Ebrahimi proposed a work to solve the problem is memory-based Experience Replay (ER) for classical catastrophic forgetting. In the quantum realm, we can leverage this technique by searching for the right set of parameters using the additional support of buffered ER samples. When augmented with incoming samples in a continual learning setup, these samples help as an ideal set of parameters that the model has to look for even when it is starting to forget previously learned samples.

With the rise of utilising quantum speedups, we try to investigate a most prominent problem that exists in deep learning, i.e., catastrophic forgetting by introducing a quantum neural network. Though, it has been studied extensively in classical literature, still most of the claims are theoretical and based on reducing the slight effect of catastrophic forgetting. In this work, we propose a novel hybrid architecture that helps reduce catastrophic forgetting. First, we compare and study how the quantum layer is affecting the neural network. For that, we plot training loss curves for CIFAR-10 [12] and MNIST [13] datasets with 30 epochs without continual learning for the classical neural network and the hybrid neural network. A hybrid neural network is the combination of classical layers with the quantum layer as a replacement of the prediction layer. Then using the same architecture, we perform continual learning by introducing more samples at the 29th epoch. The samples added are proportionate to the dataset size and split that has been taken into account while experimenting.

Then we plot the curve for training loss for both approaches (classical and hybrid). Once we confirm the hypothesis about less effect of catastrophic forgetting on hybrid neural networks, we then analyse the explanations for the regions that the hybrid network is looking for in order to remember the class-specific features in the training procedure. Finally, we discuss what are the possible mitigation strategies and compared classical neural networks, classical SVM, pure quantum neural network, hybrid SVM, and hybrid quantum neural networks (CQural).

In this research work, we start with investigating recent modern machine learning techniques that can be adapted with classical-quantum hybrid architecture. We start with investigating Transfer Learning, Multi-Task Learning, Federated Learning and Neural Architecture Search. In addition, we also discuss current challenges in each of these modern techniques and how classical-quantum hybrid architecture can overcome those limitations. Finally, we discuss about explainablity based hybrid neural network deep learning for CIFAR-10 and MNIST datasets.

*A. Motivation*

Due to the probabilistic model and energy-based computations, quantum models are highly expensive and time taking to run in the NISQ (Noisy Intermediate Scale Quantum) era. Therefore, current ML tasks that quantum computers can solve are the ones that classical neural networks used to solve 10 years ago [1]. With the increase in error-free quantum computers and multi-qubit systems, we will be able to perform massive ML tasks. We have openly discussed modern deep learning tricks where quantum variational circuits could be useful with their limitations, at the same time it is challenging to implement such architectures without sufficient knowledge of empirical proofs. At this stage, using the multi-qubit system for multi-class classification is not as robust as performing single-qubit tasks for two-class classification. There is the possibility of three such architectures: Classical-Quantum, Quantum-Classical and Quantum-Quantum. Though a wide study has been done on types of architectures in quantum information, it has not been studied how the hybrid architecture is performing in a continual learning setup. The possible direction that we have identified is to first study the effect of continual learning and perform an evaluation on the machine learning hybrid model with iteratively increasing epochs with new data samples. Additionally, a GradCAM [14] based visualization for the two-class classification model in order to investigate the regions that model is looking for in the correctly predicted samples. Previously, minuscule work has been done in the direction of visualizing the features that are most important for a class in continual learning. In addition, the idea of using saliency maps as part of the training procedures to mitigate catastrophic forgetting and visualizing the areas that are being forgotten is a new challenge.

*B. Objective of this novel work*

Catastrophic forgetting in the modern neural network at several learning paradigms has become a hurdle for the future of artificial general intelligence. The tendency of artificial neural networks to forget previously learned knowledge when presented with new knowledge is a challenge for online learning tasks. With the developments of quantum information and computation for theoretical speedups, this experiment-based research starts with a single qubit acting as a discriminator in the last layer of the classical neural network. The objective is to leverage the quantum speedup in machine learning, evaluating on the basis of forgetting statistics and proposing a novel metric to mitigate catastrophic forgetting using hybrid neural networks. Key objectives are as follows:

- Prepare a hybrid classical-quantum neural network that performs better than current deep learning models in a continual learning setup.
- To run multiple experiments and check for catastrophic forgetting when the model is tested with additional random samples from the same domain.
- Analyse and compare different models with the proposed architecture at the cost of the training procedure.
- Provide explanations and the right evidence for the correct predictions of the hybrid neural network on the CIFAR-10 and MNIST dataset.
- Methodology to mitigate catastrophic forgetting using the right evidence and activation maps in continual learning setup.

*C. Contribution*

Our research is focused on classical-quantum hybrid architecture for two-class classification. The primary goal is to understand the effect of continual learning on hybrid architecture performing evaluation metrics (precision, recall, training loss, and accuracy) on the CIFAR-10 and MNIST datasets. In addition, the secondary goal is to visualize the regions using GradCAM, that are participating in the predictions and what the model is looking for inside the correctly sampled classes. Lastly, we intend to propose mitigation techniques to decrease the effect of catastrophic forgetting using a novel metric.

Key contributions of this research work are as follows:
- Performing classical and quantum continual learning with MNIST and CIFAR-10 datasets to show the effect on catastrophic forgetting by plotting training curves for each setting.
- Novel hybrid neural network that performs two-class classification for top two classes from MNIST and CIFAR-10 datasets.
- Come up with explanations and visualizations for class-specific features using GradCAM to understand what features play important roles while learning of a hybrid neural network.
- Comparing classical CNN, SVM, Quantum Neural Network, hybrid SVM and proposed hybrid neural network architecture for MNIST and CIFAR-10 datasets.

Powerful Deep Learning models are difficult to interpret and thus often treated as black-boxes. Hence, we require a method that can interpret how the inner mechanism works inside deep learning models. There has been great research in the interpretability and explainability of artificial neural networks (for example GradCam and GradCam++, LIME model etc). Most of the explainability mechanisms work in the sense that they highlight the specific regions in the image that led to the prediction of the neural network as a piece of evidence. In a continual learning setup, the idea is to learn continuously with the changing environment such that the model adapts to the changes while learning. Knowledge is transferred from previously learned tasks and new knowledge is stored for future use prior to refining it. But, when you try to learn in a continual learning setting, the major challenge arises when your model starts to forget previously learned information or it starts to predict ambiguous classes (or semantic drift).

### D. Organization of paper

In section II we rigorously survey modern deep learning techniques that have powered artificial general intelligence. We also added a new line of research that can take place with their respective quantum alternative for each of these techniques. In section III we discuss the proposed architecture by starting with the objective and then explaining catastrophic forgetting. Then we propose a novel scheme based on quantum circuits and parameter shift optimization procedure used by the hybrid neural network. Further, we have discussed the analysis and understanding of the results from the experiments. Additionally, we have proposed a novel technique for the mitigation of catastrophic forgetting by explainable artificial intelligence techniques. Finally, we discuss the shortcomings and limitations of the proposed architecture along with a conclusion and future work.

## II. RELATED WORK

This research article is being proposed to work on recent developments in classical machine learning using the hybrid neural network. In this section, a theoretical analysis is done for each of the quantum alternatives of a modern machine learning algorithm and its applications. We study the classical techniques that make the machine learning pipeline efficient. Transfer Learning utilizes features that are learned on previous tasks in order to learn a new task. This technique is unique on its own because it reduces the redundant use of learning the same features again as we know earlier stages of the deep learning models learn basic horizontal and vertical edges. Then, we discuss Multi-Task Learning (MTL), which allows the tasks to be learned such that the features that could not be learned in isolated learning environments could be learned in a shared environment. Further, we discuss Federated Learning and Neural Architecture Search and their possible quantum alternatives.

### A. Transfer Learning

Given a source domain $Ds$, task domain $Ts$, a target domain $Dt$ and learning task $Tt$, transfer learning helps to improve the learning of target prediction function in $Dt$ using knowledge in $Ds$ and $Dt$ where both datasets are independent or both tasks are different. In simpler words, in the modern applications of machine learning, it is very convenient to start with a pre-trained model on some task and train it for a required particular task. The model becomes able to recognize and apply the knowledge that is learned previously to new tasks. There are three major issues with transfer learning that asks, What to transfer? How to transfer? And When to transfer? [15] In this proposed research, it is claimed by this article that these three questions can be answered in the quantum realm.

*1) What to transfer:* This question asks which part of knowledge can be transferred across tasks. Some features are very important for transferring knowledge, while others are not for the target domain. Using quantum importance sampling [16], we can sample only those samples and features that play a vital role in the training target domain.

*2) How to transfer:* After discovering what has to be transferred, a learning algorithm has to be devised that can transfer the knowledge. With the help of variational quantum circuits, also called QNN [17] and a classical NN, it is possible to achieve a good performance. Apart from classical-classical transfer learning, there exist three other types of transfer learning: classical-quantum (CQ), quantum-classical (QC), and quantum-quantum (QQ) [18]. The focus of this research is to study the use of classical-to-quantum networks to solve difficult tasks that are infeasible for classical models. Currently, we have powerful deep-learning models that give near-human accuracy. With a joint CQ architecture, solving high-dimensional classification tasks may become easier. In order to explain this, consider a large 1000-class classification task. Classically, we can train a powerful deep learning model on top of it to extract class-specific features to make decision boundary that generalizes well. However, using only classical will consume a lot of time to train. Using quantum variational circuits as a catalyst such that QVC or QNN [17] learns only highly confident features that are specific to a class, in other words, using the classical model to learn broad features (that include horizontal and vertical edges) and quantum model to learn class-specific features (that includes sharp features of a particular class). To learn broad features with QVC, it will require a large number of qubits to handle the huge demand of learning elements [19]. With the current limitations of quantum computation, it is nearly impossible to train a large-scale QQ transfer learning paradigm.

*3) When to transfer:* In transfer learning, domains should be structurally similar to exploiting prior knowledge. It is of no use if the model has learned to classify cars and uses its knowledge to learn the classification of flowers. In other words, when classification task is of similar domains, it is empirically proved that transfer learning performs better. In a CQ setup, it becomes a challenge to identify at what threshold it is required to shift learning from the classical model to quantum. Using deep visualizations and saliency maps may help to identify the stopping criteria for a particular task [20].

A deep learning model should not forget previously learned knowledge. Starting layers tend to learn vertical and horizontal edges that are generic. Intermediate layers start to learn the combination of these horizontal and vertical edges. The final layers of the neural network model learn class-specific features. Stopping criteria for CQ transfer learning methodology can be after the intermediate layers such that all the layers that are classical are frozen.

### B. Multi-Task Learning

Multi-Task Learning (MTL) [21] has proved a remarkable use in classical machine learning. For example, in image captioning task, MTL has proved a state-of-the-art performance. For an input space (X), output space (Y), the probability distribution (D), and training set (S) there exist a set of tasks ($T_1$, $T_2$, ..., $T_k$) and the goal is to find a good classifier for each task as shown in the Fig. 2.

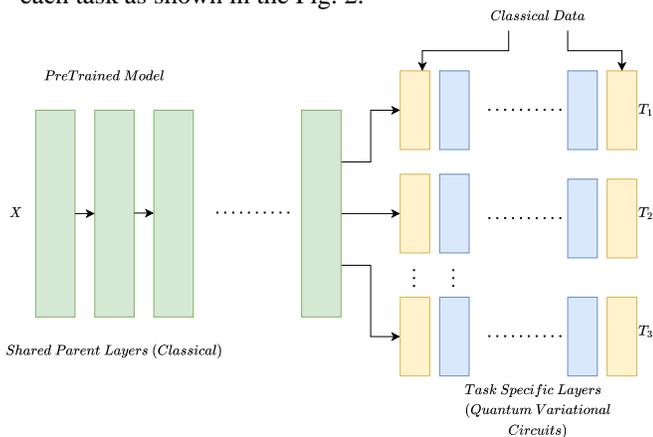

Figure 2 Quantum Multi-Task Learning

In classical MTL, the deep learning architecture is divided into two parts. Shared layers and task-specific layers. The division is because of the fact that early layers learn almost the same features. However, it is because of the last layers, the model becomes task-specific. In classical MTL, a single shared layer is learned for different tasks, however, in later stages, individual losses will update individual layers (also called task-specific layers) while the parent layer (shared layer) updates on all individuals [22]. In this technique, we can follow the CQ methodology to learn shared layers classically while task-specific layers as QVCs or QNNs. This kind of technique performs well when tasks are similar in nature and could not be learned in isolation. Classical-Quantum MTL is expected to utilize all the pre-processing of the huge classical data and efficiently learn class-specific features at the task-specific layers.

### C. Federated Learning

The primary objective is to learn a hybrid QNN for a real-world setting to perform image classification. Once learned a global classifier, takes updates from multiple quantum computers residing at the edge it speeds up the federated learning because of the quantum processing. In addition, the objective is to study the quantum alternative to the Federated SGD process and Federated averaging which is an efficient way to perform classical FedML. In the paper "Federated Quantum Machine Learning" [23] the author has addressed global averaging for aggregating the client data. Previous research has shown FedSGD is helpful when a client has limited computing power, however, in the quantum world we have given exponential speedup in the hands of the client, therefore, global averaging will work much faster. This way communication between the server and client can be reduced. Instead of gradients weights are transferred back to the server. And the server performs the aggregation for all the client weights. This is repeated for multiple rounds, however, the number of rounds in total reduces because weights are getting transferred in mini-batch.

### D. Neural Architecture Search

One of the popular techniques in modern deep learning is neural architecture search (NAS) also called AutoML. Deep learning is an extremely useful technology that has proved human-level accuracies and even more than that (in time-limited humans). However, it requires a human-machine learning expert to pre-process data, select & engineer features, select a model from the model family, and then optimize the hyperparameters. In order to mitigate this challenge, AutoML provides methods and processes to make ML available from lab to real life to improve the efficiency of ML and accelerate research on ML. AutoML helps to quickly and cheaply deploy ML as a black box in their work area and use results out of it. Using AutoML it is possible to automate hyper-parameters, selection of the best architecture and model, data pre-processing, and feature selection [24]. Classically, DARTS and one-shot NAS have given good theoretical bounds to search for the best architecture. It works on the principle of weight sharing among different small networks that reside under a parent network. Evaluating each of the child network and then coming up with the final network is resource-intensive and time-consuming. The same approach can also work in training the entire network at once by learning multiple models in a single epoch.

NAS can be imported into the quantum world with two fashions. Quantum NAS (classical-quantum) and Quantum Architecture Search (quantum-quantum) [25]. There has been little research on the QQ variant of NAS or QAS. However, in this particular section, we look at the hybrid model of NAS and understand how quantum speedup can help in classical NAS. QAS can be proposed as a classical-quantum hybrid way of finding the best deep learning setting by applying search. Among the tasks completed in order to find the best neural architecture, pre-processing of data and feature selection can be done classically whereas quantum speedup can bring a significant change in searching for the best deep learning model among the pool of available models using quantum alternatives of genetic and evolutionary algorithms [26].

## III. Proposed architecture for Quantum Continual Learning

As we already discussed why continual learning is important for artificial general intelligence, we envisage that introducing a quantum neural network as a hybrid learning approach will make it robust to catastrophic forgetting. This is because of the fact that quantum neural networks perform well in finding the difficult features in the feature space that are far off from the decision boundary [27]. In simpler words, quantum variational circuits are good in exploring complete feature space. To avoid catastrophic forgetting in continual learning it is necessary to remember class-specific features, therefor we explore the field of quantum continual learning and further try to explain it by deep visualizations. Details about our proposed approach are described below:

### A. Catastrophic Forgetting

Examples that tend to be forgotten by the learning model in later stages of the training process, in other words, the predictions which were correct in the earlier cycles of learning are labeled incorrectly in the later stages of the learning process bringing a shift. Such a shift in the prediction is called forgetting. In addition, the examples which cannot be forgotten even in the later stages of the learning process are unforgettable. The informativeness is judged on the basis of examples that are forgettable. That is we need to identify that the examples which are easily forgotten will be a good candidates for generalization. Additionally, we have examples that are unforgettable throughout the training process and are good candidates to be the candidate to remember class-specific features from the dataset because of being highly informative [8]. In the addition, it is observed that with the continual learning pipeline, most of the examples are getting different labels with subsequent learning cycles. Therefore, its a need to devise an optimization function or a label-centric cost function with visualizations that can modify the continual learning pipeline so that we could identify the level of forgetting being occurred [9].

The forgettable examples are characterized by their different feature sets or underlying representations which are distant from the features of the original task samples in the input space. Precisely authors of [8], define forgettable events as when an example is classified correctly at a time (t) in one batch cycle, gets incorrectly classified at a time $(t + \mu)$ where $\mu$ is the time spent after the mini-batch. They hypothesized and then revealed that the examples which are consistently forgotten do not have anything common in feature space with the tasks which are classified correctly for that task. In addition, this helped in two ways. First, in reducing the dataset size without losing generalization capabilities. Secondly, analyzing the forgetting examples to check which sample is the most informative and which samples are least informative. Informative means that, from which example you can extract most of the features which could help better generalize the task. This helps in eliminating those examples which are of no use. In simpler words, the examples having different features which lie far from the decision boundary but belong to the classification task, are the most informative and important to generalize whereas the ones which are classified with the best confidence score are the least informative. Further characterized the forgetting events on the basis of (1) stability across seeds where they concluded that small forgetting brings confidence towards the correct classification of that example, (2) forgetting by chance, (3) first learning events, where some examples which are learned in the earlier part of the training tend to be forgettable and unforgettable examples, are learned in very later stages of the training process (4) misclassification margin that is the difference between chances of getting classified as an original label and chances of getting classified as other class(es) label. Lastly, previous work [8] proves experimentally that the removal of unforgettable examples does not disturb the overall performance of the training accuracy and generalization capabilities.

Another recent interesting research [9], says that the more predicted label fluctuations you see, the higher the label dispersion will be and vice versa. Hence, recall we mentioned that when the model starts to shift its predictions in later stages of the training, we say that it has undergone catastrophic forgetting. Therefore, in active learning when we equip this kind of acquisition function authors have obtained benchmark results. By definition, AL is a learning technique that helps researchers and practitioners to start training the model with limited labeled data. This technique achieves significant performance when it is allowed to choose which sample to label. Whatever labeled pool you have, start with that. Afterward, allow the model to sample from an unlabelled pool using the acquisition function. Once labeled, the newly classified example is then sent to the labeled pool and re-train using an updated training set. This process is repeated till the budget exhausts [28].

### B. Methodology

*1) Defining Quantum Circuit:* A standard quantum circuit is used for the experiment. A quantum circuit [29] is a series of gates that are applied to perform qubit transformations. A quick recall of all quantum gates will be helpful before reading further text. A combination of Hadamard gate [30] and single-qubit rotation over the y-axis shown in Fig. 3, is applied to the qubit, in the quantum circuit definition. It is easier to take analog matrices and matrix transformations to understand what's happening in the quantum circuit. The *run* function calculates the probabilistic value of the states and returns the most expected value of $\pi$ rotation [2].

*2) Hybrid Function:* The curated hybrid function leverages the optimization power of classical as well as quantum computers. It is expected that classical functions help to optimize and find out precise features from the dataset, however, quantum functions are expected to learn features beyond classical functions [31]. Hence, it is efficient to add a quantum layer at the end of the classical architecture because the hybrid layer [27] finds good probability values. One reason to work on classical-quantum architecture with backpropagation is not efficient with current quantum architecture based on the literature on huge machine learning datasets, as it would create a huge requirement of qubits at starting layers of the network to take inputs. In addition, no theoretical bounds state that we can perform quantum backpropagation as robust as classical one [32]. In simpler words, for example, considering image datasets, if we want to inculcate a fully quantum-based neural network for such datasets, it is unrealistic with current technology.

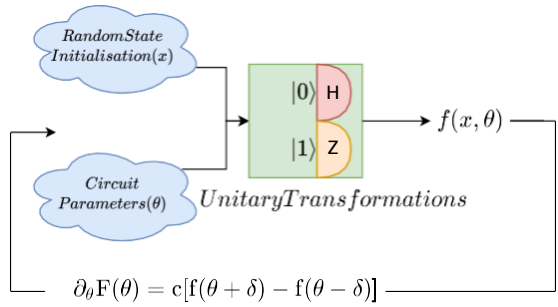

Fig. 3. Hybrid Function using Parameter Shift Method

*3) Hybrid Neural Network (CQural):* We import CIFAR-10 and MNIST datasets and limit the size of the dataset to the top two classes for our analysis. To keep the training faster, we have just used 100 samples from each class. In the neural network architecture, we have used two convolutional layers, one dropout, two fully connected, and one hybrid respectively as shown in Fig. 1. Two convolutional layers are used that apply 2D convolution over input signals. A 32X32 CIFAR-10 image is taken as the input signal passes *conv*1 results a 28X28 feature map which then passes *conv*2 resulting in a 24X24 feature map. A simple *dropout*2D layer is then used that randomly drops the connections. Further 2 fully connected layers are used to map like a simple neural network that applies linear transformation. Once we have classical information from the *fc*5 we will transform the classical data to a quantum state using Amplitude Embedding [33]. Quantum information is passed to the quantum layer that applies unitary transformations and returns back classical data that is further used as classification probabilities. Specifically, an Adam optimizer with a learning rate of 0.001 is used based on the experimental analysis. The adaptive Momentum Estimation optimizer has an adaptive learning rate that prevents the optimization procedure not to oscillate in space. This is done to reduce the oscillations of the gradient in the optimization procedure. In contrast, RMS Prop requires more steps to converge, and, Adam optimizer will make sure the convergence happens faster and reach optima [34]. The loss function used is negative log-likelihood loss (NLL Loss). NLL Loss adaptively adjusts the learning such that for minimum and maximum parameters it will take big steps and cautious steps respectively [35].

*4) Training:* We have tested the experiment 5 times with running cycles multiple of 5 in an increasing fashion. In addition, as part of continual learning, we inflated a new dataset in the middle of the epoch between 28 to 30 and found training loss overshoot sharply. Observations and results shown in Fig. 5 and Fig. 6 from the training are discussed in the next section. There is no addition of a quantum optimizer in the training procedure. We have used the same *model*() function to get the output and calculate the loss. Losses from each of the cycles are stored and analyzed. A detailed learning procedure of hybrid training is shown in Fig. 4. We start with building a hybrid neural network architecture using classical neural networks and quantum variational circuits. The circuit is initialized with random parameters that adjust with the optimization function called as parameter shift rule [27]. The prediction probabilities are given by the classically converted optimized quantum states.

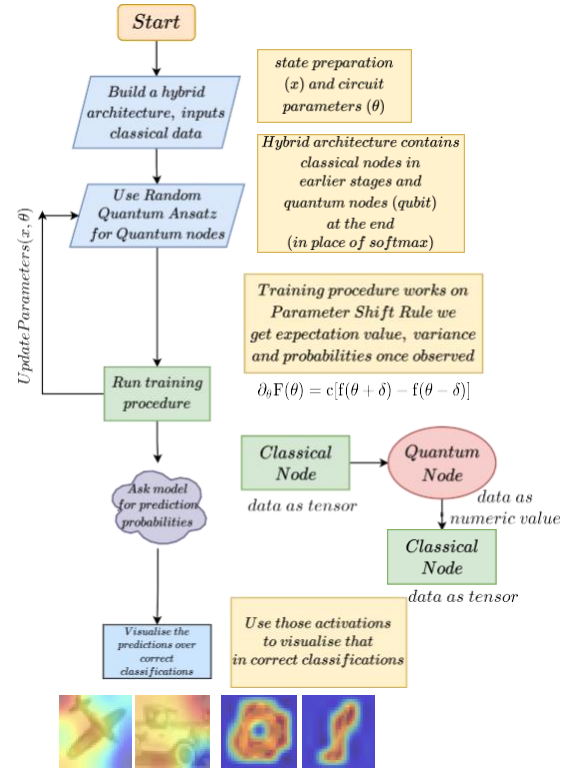

Fig. 4. Algorithm Explained.

## C. Mitigating catastrophic forgetting

Starting from the previous line of work, we started investigating Regularization based Model Distillation (MD) [11] and Memory based Experience Replay (ER) [36]. And came up with a hybrid model such that we can leverage the functionalities of both of techniques in a combined form. ER draw samples from each task kept into the task pool and the model can revisit the pool as it learns previous tasks up to a threshold accuracy. In order to know how deep visualizations and explanations can help the time-limited humans to get the explanation for the enhancement of experience replay for the model. In order to refine the training process, the training samples are added to store the predicted model explanations. This will help the model to memorize the locations accurately when asked for explanations and where the model paid attention. Saliency maps (GradCAM++ and LIME with gradient boosting) are used such that it highlights the specific regions that are above 90% confidence for the predictions of training samples. Models that work well with such kind of setting are used. The core idea is when you add your saliency maps and explanations to your learning procedure and encouraging the model to remember the explanations and pieces of evidence helps in long-term learning in a continual learning setup. In other words, when you are learning in a continual learning setup, once you learn a specific modality of tasks, feed the model with its explanations and spin your learning procedure again such that your model now memorizes the explanations along with the training weights. This boosts the decision-making power and classification accuracy of the model for the long term even when you add more tasks. In order to reduce the learning cycles, empirically it is suggested that the gradient maps and explanations or the GradCAM outputs should be learned with new task procedures.

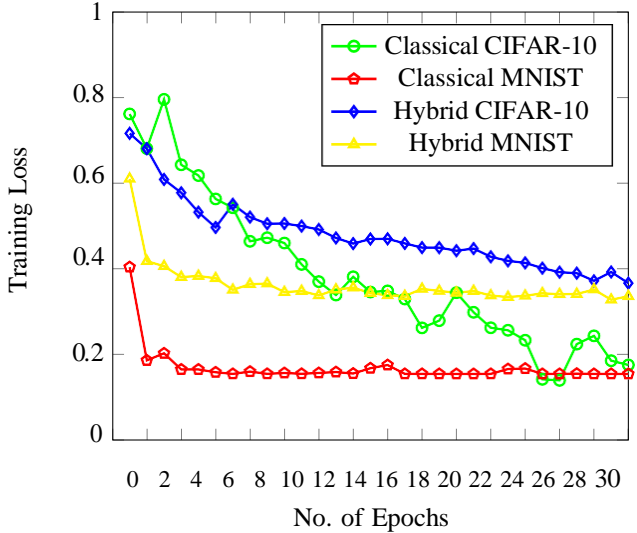

Fig. 5. Training Loss curves without continual learning. This curve shows a comparison between classical and hybrid architectures. We can observe that hybrid architecture and classical architecture give the same level of accuracy for the CIFAR-10 and MNIST datasets.

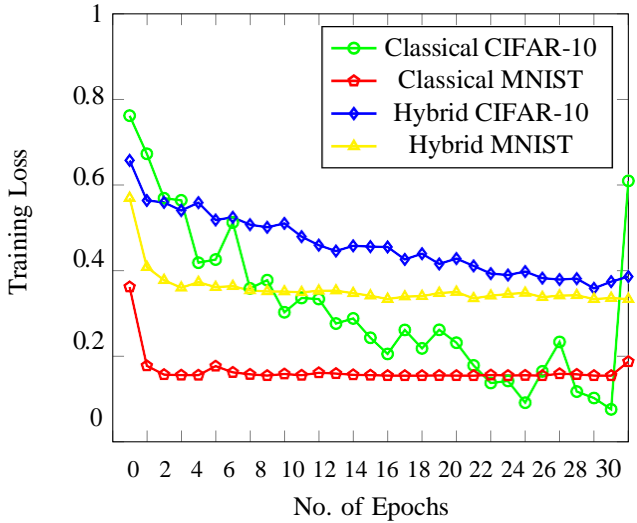

Fig. 6. Training Loss curves with continual learning. The graph for Hybrid architecture does not show multiple evidence of forgetting whereas classical models have been consistent in showing the rise in training loss between multiple epochs.

### D. Novelty

The contribution of this work will add a robust mechanism to the current continual learning procedures. The models that tend to forget quickly in several training cycles in the continuous flow of input data, this model proves to work better than existing low-computation-based architecture. Not only this, it will be easy to analyze which features are being forgotten and which features are participating in class-specific decision-making. Even if the context of new inputs remains the same, current neural networks tend to shift their decision boundary drastically, where our model withstands the forgetting events at the same time while remembering the class-specific features even if they are far off in the feature space. Innovation factors include a novel hybrid architecture that comprises CNN and a quantum layer. This hybrid network is then used for continual machine learning where it provides an efficient way to improve correctness and reliability.

TABLE I
CLASSICAL MODEL - CIFAR-10 50:50 SPLIT (PRECISION AND RECALL FOR AIRPLANE)

| Epoch # | Cross Entropy Loss | Accuracy (%) | Precision | Recall |
|---|---|---|---|---|
| 5 | 0.5492 | 72.0 | 0.60 | 0.37 |
| 10 | 0.6155 | 72.0 | 0.54 | 0.37 |
| 15 | 0.3648 | 82.5 | 0.50 | 0.54 |
| 20 | 0.3869 | 84.0 | 0.53 | 0.50 |
| 25 | 0.4154 | 81.0 | 0.48 | 0.44 |

TABLE II
HYBRID MODEL - CIFAR-10 50:50 SPLIT (PRECISION AND RECALL FOR AIRPLANE)

| Epoch # | NLL Loss | Accuracy (%) | Precision | Recall |
|---|---|---|---|---|
| 5 | 0.6330 | 70.5 | 0.54 | 0.37 |
| 10 | 0.5258 | 51.5 | 0.45 | 0.45 |
| 15 | 0.7460 | 80.5 | 0.54 | 0.64 |
| 20 | 0.8204 | 82.5 | 0.51 | 0.64 |
| 25 | 0.8410 | 84.0 | 0.51 | 0.64 |

TABLE III
CLASSICAL MODEL - MNIST 50:50 SPLIT (PRECISION AND RECALL FOR ZERO)

| Epoch # | Cross Entropy Loss | Accuracy (%) | Precision | Recall |
|---|---|---|---|---|
| 5 | 0.161 | 99.5 | 0.56 | 0.55 |
| 10 | 0.1595 | 100 | 0.52 | 0.52 |
| 15 | 0.1567 | 100 | 0.44 | 0.44 |
| 20 | 0.1605 | 100 | 0.55 | 0.55 |
| 25 | 0.1609 | 99.5 | 0.45 | 0.45 |

TABLE IV
HYBRID MODEL - MNIST 50:50 SPLIT (PRECISION AND RECALL FOR ZERO)

| Epoch # | NLL Loss | Accuracy (%) | Precision | Recall |
|---|---|---|---|---|
| 5 | 0.7353 | 71.0 | 0.50 | 0.79 |
| 10 | 0.7361 | 73.0 | 0.49 | 0.78 |
| 15 | 0.7337 | 72.5 | 0.51 | 0.76 |
| 20 | 0.7343 | 73.5 | 0.52 | 0.81 |
| 25 | 0.7438 | 75.0 | 0.50 | 0.77 |

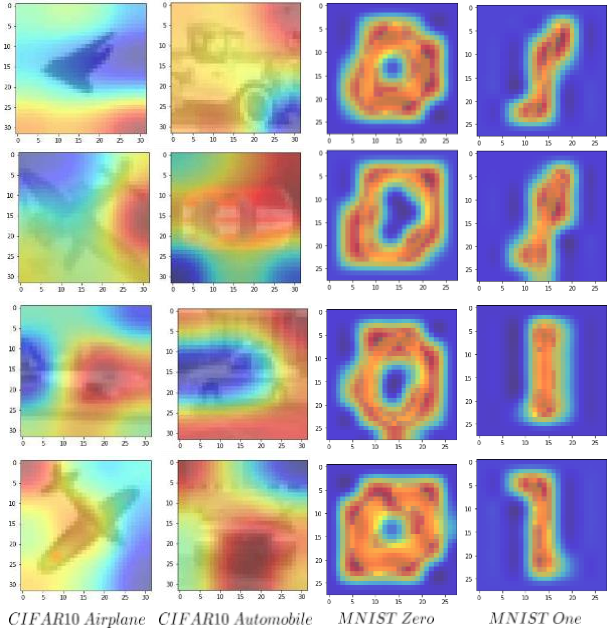

Fig. 7. GradCAM visualization for CIFAR-10 and MNIST

## IV. RESULT AND ANALYSIS

We have used state of the art approaches and implemented our own architecture to compare and analyse the classical, quantum and classical-quantum deep learning models.

### A. Implementation Environment

We have used Google Colab Pro to run python 3 notebooks. We used T4 GPU with 25 GB of RAM size. IBM qiskit (0.34.1 version) library is used as a quantum simulator with its additional requirements. PyTorch toolbox is used for the processing of the dataset and learning procedure. In addition, tensorflow is used to slice the neural network layers to extract features and perform deep visualizations for CIFAR-10 and MNIST datasets. We also limit our dataset to two classes with equal ratio (50:50) of the samples to prevent for biasness. The training and testing split is 80:20 respectively (we found almost same results for 70:30 split). We have set qasm-simulator as simulation environment for the quantum circuit backend as

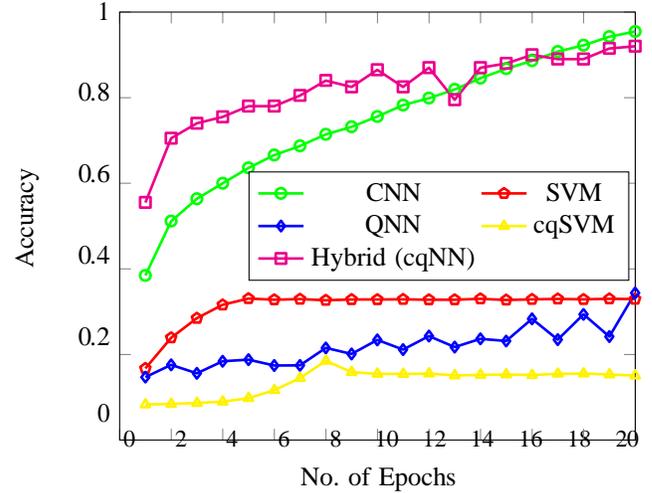

Fig. 8. Comparing four different models with the proposed architecture for CIFAR-10 dataset. CNN and Hybrid cqNN gives almost same accuracy however, remaining classifiers give sub-optimal results

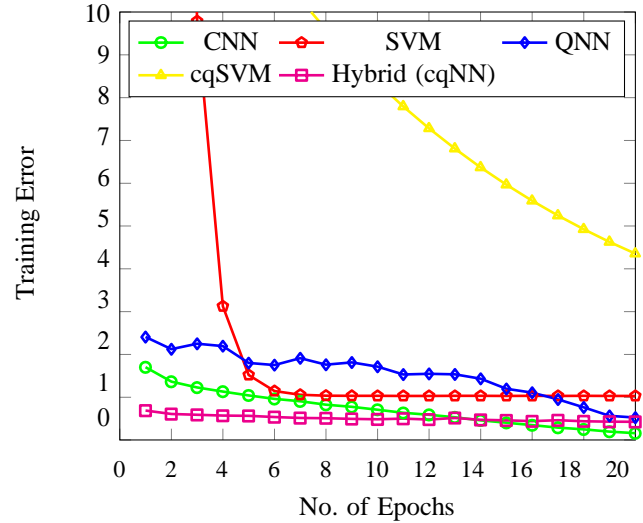

Fig. 9. Comparing four different models with the proposed architecture for MNIST dataset. Error for CNN and cqNN is almost same that justifies cqNN is as good as CNN

it provides all standard functions for qubit operations as a quantum simulator.

### B. Performance Measures

To measure the performance of test data we analysed training loss and test accuracy. Sklearn library is used to produce classification reports, confusion matrices, ROC curves, and precision-recall curves. Classification report helps to provide a top view of the results and major classification metrics in text output. Additionally, to analyse the performance of the classification model, we study the confusion matrix on the test dataset. ROC Curve is a simpler measure of confusion matrices. The Receiver Operator Characteristics curve is a simple way to summarize. The vertical axis of the graph is sensitivity (True Positive Rate) and the x-axis is (1-specificity)

False Positive Rate with varying thresholds. The threshold is a kind of decision boundary that enables the samples to classify into their respective classes. Precision is the number of times the class got correctly classified divided by the number of times the model assumed it was that class. However, Recall is a number of times model classified it was that class divided by the number of times it was actually that class. In our results, hybrid neural network performs as good as a classical CNN network in Fig. 8 and Fig. 9, however, CNNs are prone to catastrophic forgetting as we have seen earlier in Fig. 6.

*C. Visualizations for Explainability*

The visualizations shown in Fig. 7 are for the CIFAR-10 and MNIST datasets. As you can see, our hybrid model has learned features so nicely for the MNIST dataset, however, for the CIFAR-10 dataset, it is still struggling to learn the features. However, the model has learned the best possible features from each of the datasets. This new line of work actually opens a new direction to work on quantum continual learning such that the hybrid model overcomes the fundamental problem of classical deep learning models that tend to forget over a period of time.

Two key takeaways from the quantum continual learning are as follows:
1) Quantum continual learning is more robust than classical continual learning as we have seen through the visualizations and metrics. A hybrid model is able to learn class-specific features very precisely than the classical model. It is because the quantum circuit is actually performing a search task over the features and doing the importance sampling for those features.
2) With the increasing epochs, the hybrid model has shown a good amount of accuracy (Table II and IV ) however, the native belief system of the features remains the same that can be shown via GradCAM visualizations that show that hybrid models are able to identify the class-specific features from early stages of training.

*D. Comparison Analysis*

Proposed hybrid architecture performs provably well for the continual learning setup as we can observe that when the learning procedure is introduced with some new random data samples, the loss overshoots for classical model comparatively to hybrid model after 25 epochs. On comparing classical, quantum and hybrid approach in Fig. 8 and Fig. 9, we can say that though current neural networks perform well than handcrafted hybrid architectures, however, hybrid architectures perform exceptionally well under continual learning setup shown in Fig. 5 and Fig. 6 with Tables I, II, III and IV. We investigated and found that when the learning procedure is introduced with new samples (the samples that are not part of existing dataset) the training loss gives a sharp rise while working with hybrid architecture it is not the case. This proves and justifies that proposed classical-quantum hybrid architecture is superior at learning class-specific features more easily than classical one. Appreciating this will help us to understand the explanations shown in Fig. 7 as well.

*E. Shortcomings and limitations*

The major limitation of this research work is that it is based on 1-qubit hybrid architecture. However, the objective of this research was to understand the performance metrics and visualizations for explainability in a continual learning setup. This research can be extended to multi-qubit multi-class architecture. The proposed neural network can be extended with the same in a classical-quantum manner. Another limitation of this work is that it is purely based on experiments and the theory is developed on empirical proofs. Therefore, it is possible while extending this work to multi-qubit multi-class architecture, results look different than presented in the paper. One can extend this work by performing the following steps:
1) Working on quantum-classical and Quantum-Quantum architectures and reproducing same evaluation metrics and visualizations to compare which architecture has lowest catastrophic forgetting in continual learning setup.
2) Use multi-qubit multi-class routines by changing the neural network layers.
3) One can extend this work for other image datasets that have complex feature space in order to investigate how hybrid architecture learns the classically non-traceable features.
4) Hybrid model takes four to five times more time to learn. However, it has better explainability. In addition, it is expected to reduce the learning time with an increased number of qubits in a hybrid model.

V. CONCLUSION AND FUTURE DIRECTION OF RESEARCH

In this paper, we have proposed a novel hybrid architecture that combines classical and quantum properties of solving deep machine learning tasks. We started with investigating catastrophic forgetting in continual learning and analysed how classical models fail when it comes to solving new tasks with time. Whereas, we found hybrid architectures empirically performed well with such tasks with an insignificant amount of forgetting. To support this claim we also analysed the regions as part of explainability in order to understand the features that the hybrid model is looking for. We found that hybrid models actually look for the features that are class-specific and remember them even when those features are far from the decision boundary in feature space. In addition, we propose a novel technique for minimizing the catastrophic forgetting using the saliency maps of correctly predicted classes as part of the training process in a continual learning setup that makes the deep learning model look for the right pieces of evidence in new data samples. Finally, we compared classical, quantum and hybrid machine learning models on MNSIT and CIFAR-10 datasets. And found that hybrid architecture does not show catastrophic forgetting during the continual learning. Full code is available here https://github.com/s4nyam/cqural

A new way of solving traditional machine-learning problems is envisioned for future work. Solving tasks that take considerable time to learn and process, are expected to be solved under quantum optimization procedures. This research paper proposes to work on parallel problems of modern machine learning techniques including Transfer Learning, Multi-Task Learning, Federated Learning, and Neural Architecture Search.

ACKNOWLEDGEMENT

This work is partially supported by Indian Institute of Technology (IIT), Jodhpur, Rajasthan, India and DST India.

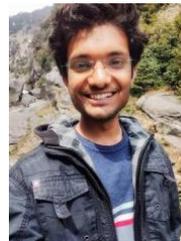
Sanyam Jain received his Bachelor of Technology degree from University of Petroleum and Energy Studies, Dehradun, India. He is a Junior Research Fellow at Indian Institute of Technology, Jodhpur from 2020 in Interdisciplinary Quantum Information and Computation. His research interests include deep learning, quantum machine learning and explainable AI.